\documentclass[twoside]{article}

\usepackage{aistats2023}

\usepackage{soul}
\usepackage{graphicx}
\usepackage{xspace}
\usepackage{xcolor}
\usepackage{caption}
\usepackage{subcaption}
\usepackage{xr}

\usepackage{url}
\graphicspath{
    {figures/}
}

\newcommand{\paragraphb}[1]{\noindent{\bf #1.}}

\newcommand{\numfolds}{10\xspace}

\begin{document}

%

%

\twocolumn[

\aistatstitle{DeepG2P: Fusing Multi-Modal Data to Improve Crop Production}

\aistatsauthor{ 
 Swati Sharma\textsuperscript{a}\And Aditi Partap\textsuperscript{b}\And Maria Angels de Luis Balaguer\textsuperscript{a}\And Sara Malvar\textsuperscript{c}\And Ranveer Chandra\textsuperscript{a} }

\aistatsaddress{
\textbf{a} Microsoft Research, Redmond\And \textbf{b} University of Stanford \And \textbf{c} Microsoft Research, Brazil} 
]

\begin{abstract}
   Agriculture is at the heart of the solution to achieve sustainability in feeding the world population, but advancing our understanding on how agricultural output responds to climatic variability is still needed.  Precision Agriculture (PA), which is a management strategy that uses technology such as remote sensing, Geographical Information System (GIS), and machine learning for decision making in the field, has emerged as a promising approach to enhance crop production, increase yield, and reduce water and nutrient losses and environmental impacts. In this context, multiple models to predict agricultural phenotypes, such as crop yield, from genomics (G), environment (E), weather and soil, and field management practices (M) have been developed. These models have traditionally been based on mechanistic or statistical approaches. However, AI approaches are intrinsically well-suited to model complex interactions and have more recently been developed, outperforming classical methods.   
   Here, we present a Natural Language Processing (NLP)-based neural network architecture to process the G, E and M inputs and their interactions. We show that by modeling DNA as natural language, our approach performs better than previous approaches when tested for new environments and similarly to other approaches for unseen seed varieties. 

\end{abstract}

\section{INTRODUCTION}

The ability of agriculture to feed growing populations has long been a source of alarm and continues to be high on the global policy agenda. According to the Food and Agriculture Organization (FAO) of the United Nations (UN) report on the State of Food Security and Nutrition in the World in 2022~\cite{fao_repurposing_2022}, food security deteriorated in 2021. Only exclusive breastfeeding among babies under six months of age and child stunting have made progress toward the 2030 global nutrition objectives, while anaemia among women and adult obesity are rising. Updated estimates suggest that healthy and nutritious diets are unaffordable for almost 3.1 billion people around the world. UN organizations predict that population increase will continue, and that by 2050, it will reach between 8.3 and 10.9 billion people. According to experts, such growth rates would require an increase in food supply of at least 50\%, and in certain cases up to 75\%. Achieving food security needs policy and investment reforms on multiple fronts including human resources, rural infrastructure, water management and agricultural research \cite{rosegrant_global_2003}. In addition to the growing population, climate change, extreme weather conditions and industrialization have become serious challenges to global food security and agricultural sustainability~\cite{jakhar_nano-fertilizers_2022}. Meeting rising food demand in the context of global warming requires a better understanding of the climatic conditions driving food production. In this context, it is critical to investigate how agricultural output responds to climatic variability and extremes~\cite{lesk_influence_2016}. 

There is a growing body of literature that emphasizes the significance of employing emerging technology in what is called precision agriculture (PA) ~\cite{delgado_big_2019,berry_precision_2003,pierpaoli_drivers_2013}. PA is a management strategy that uses a suite of advanced information, communication and data analysis techniques, remote sensing, Geographical Information System (GIS), and machine learning in the decision making process. It helps enhancing crop production, increasing yield, reducing water and nutrient losses and environmental impacts \cite{khanal_overview_2017,aubert_it_2012,bongiovanni_precision_2004,zhang_precision_2002}.

Although imagery and crop monitoring techniques are at the heart of PA, other data types have been gaining a presence in PA. With sequencing prices getting more affordable over the years, genomics datasets are becoming another fundamental aspect that can advance PA. Agricultural genomic datasets typically consist of deep sequencing or genotyping by sequencing performed on a panel of cultivars. These data allow the identification of genetic variations across the panel and in turn allow the association of genetic variants and traits measured in the field. Genomic data has been traditionally used by plant breeders to perform genomic selection of desirable traits. Animal and plant breeders have adopted statistical methods to select candidates and accelerate breeding cycles ~\cite{meuwissen2016genomic,heffner2009genomic}. However, phenotypic variance is affected not only by genotypic components, but also by environmental components and genotype-by-environment interactions. Accordingly, prediction of phenotypes from multi-modal datasets consisting of environmental (E), including weather and soil quality information, genetic (G) and field management (M) has gained popularity ~\cite{washburn2021predicting,kick2022yield,rogers2021importance,cooper2021tackling,li2021integrated,jarquin2021utility,technow2015integrating}.

Traditionally, mechanistic models such as crop growth models have been used to predict phenotypes from multi-modal datasets~\cite{keating2003overview}. More recently, some approaches have also been developed for calibrating these models automatically with phenotypic data~\cite{holzworth2014apsim,holzworth2018apsim}. These models have evolved into a framework that incorporates many of the required physiological equations to explore changes in agricultural landscapes. Although they can achieve high accuracy, they require high quality data to estimate the model parameters for every cultivar, which makes them unfeasible for large population panels. Geneticists and breeders have also used statistical methods, such as Best Linear Unbiased Prediction (Blup), to predict phenotypes from genomic data or a combination of genomic and environmental components ~\cite{henderson1975best,crossa2013genomic,gaffney2015industry}. Although these methods have advanced our knowledge on phenotype prediction, they are limited by the linear relationship assumption among the features. AI approaches can overcome this limitation due to their ability to model complex non-linear interactions, and have more recently started to receive attention in the genome-to-phenome prediction space  ~\cite{jubair2021gptransformer,liu2019phenotype,zeng2021g2pdeep}.

Deep learning approaches have also been developed for phenotype prediction from a genetic and environmental data combination ~\cite{washburn2021predicting,kick2022yield,westhues2021learnmet}. These approaches typically involve getting a flattened embedding for genome and environment features as well as the interactions between the two, and finally concatenating the resulting embeddings. Different types of interactions have been utilized, for example, ~\cite{Washburn2021-gem-cnn} uses interactions between genome and field management data. Although deep learning approaches have shown promising results in predicting phenotypic values from datasets involving genomics, these data present some challenges due to their structure. Specifically, these datasets typically consist of tens to hundreds of thousands of genetic variants, or Single Nucleotide Polimorphisms (SNPs) (features) 
over a population panel of few hundreds to few thousand of accessions (observations). The high ratio of features to observations often requires reducing the dimensionality of the feature space prior to the application of the models. For instance, some studies have used PCA to perform dimensionality reduction of the SNPs~\cite{washburn2021predicting,kick2022yield}. However, dimensionality reduction techniques result in traceability loss of the genetic variants, therefore inhibiting the use of the model findings for any breeding application. Therefore pre-selection of SNPs as opposed to dimensionality reduction is desirable.

In this study, we aimed at using genetic, field management, soil quality and weather data to predict yield. For this, we developed DeepG2P, a deep neural network-based approach employing multi-modal fusion, that specifically captures genome-to-environment and genome-to-genome interactions. Our contributions are as follows: 
\begin{enumerate}

\item \textbf{Interpretable genomic variant selection}: Unlike previous studies \cite{washburn2021predicting,kick2022yield}, which used a PCA on the genomics feature space for dimensionality reduction, we made use of a two step process involving LightGBM followed by mutual information, to reduce the number of SNPs used to predict the phenotype. This allows us to determine the top most informative variants prior to predicting the phenotype. Knowing which variants are related to a phenotype is very desired by plant breeders and geneticists.  
\item \textbf{Genome-to-environment interaction modeling}: Using a multi-modal cross-attention to module to model genome-to-environment interactions allows us to understand how important each timestep of the weather feature is with respect to each genetic variant. To the best of our knowledge, this is the first time cross-attention has been used in this way.     
\item \textbf{Genome-to-genome interaction modeling}: We expanded the genomics input data by incorporating the genetic variant context and positional encoding. This allowed us to treat DNA as natural language, and therefore utilize NLP techniques. These approaches help us in aggregating context information from neighboring sequence. To our knowledge, previous models have not used the SNP sequence context nor NLP to address this problem. Our results show that our approach is able to outperform previous studies for unseen environments and performs similarly for unseen seed hybrids. 
\end{enumerate}
This paper is organized as follows. In section~\ref{sec:background}, we discuss the background, followed by a description of the dataset and the proposed architecture  in sections~\ref{sec:dataset} and ~\ref{sec:framework}. We end the paper with our results in section~\ref{sec:eval} and a discussion about our findings and future work in section~\ref{sec:discussion}.

\section{Background}
\label{sec:background}
We aim to understand the effect of G, E, M factors on crop phenotypes. Crop phenotypes refer to any observable characteristics of the crops such as total yield, height of the plants, or moisture content in the grain. 

\paragraphb{G} (Genome/Genotype) refers to the genetic information (DNA\footnote{DNA is a double helix composed of 2 sequences of nucleotides/bases (A,C,G,T), held together by bonds between complementary bases}) of the seeds planted. The entire genome of an organism is a very long  DNA sequence. For instance, the maize genome is ~2.9 billion bases long. 
After sequencing a panel of cultivars (crop lines included in the study) either by deep sequencing or by genotyping by sequencing, the genome sequences across the panel can be compared to a reference sequence. The small changes found in each genome with respect to the reference are typically recorded in a VCF file. These changes, called  single nucleotide polymorphisms (SNPs), can then be used to find associations between them and a phenotype, which can eventually lead to finding the parts of the DNA (genes) that affect the phenotype ~\cite{corn-g2p-gwas1}. The SNP coordinates in the VCF files are its chromosome number and its base pair position within the chromosome. Each SNP is defined by the variation that it encodes in its position. For example, a SNP may replace cytosine (C) with the nucleotide thymine (T) at a certain position of the genome. Although other types of variants exist, the genomics data we consider here, SNPs, are the most common type of genetic variation with respect to a reference genome. 

\paragraphb{E} (Environment) refers to environmental conditions such as weather and soil. Weather includes factors such as rainfall and day length, which affect the amount of sunlight and water that the crops get, eventually affecting its overall growth and phenotypes, while nutrients in soil such as nitrogen and potassium, are essential for crop growth. These data are collected by deploying weather stations and sensors in soil or collecting soil samples. More specifically, the weather stations record air temperature, humidity, solar radiation, rainfall, wind speed and direction, soil temperature, and soil moisture every 30 minutes for the duration of the growing season, while soil samples are submitted for nutrient and texture analysis to a central soil testing lab. The goal of soil sampling in PA is to assess crop nutrient needs, such that application of fertilizers can be done on a need basis. The soil data here, however, goes one step further as it is directly used to predict the phenotype.     

 \paragraphb{M} (Management) refers to field management practices that deal with soil quality and nutrient management through fertilizer application, and practices such as tilling and irrigation. It also includes weed and pest management, as well as plant disease management. The choice of management practices influence the health of the crop and the long term agricultural output from a farm~\cite{field_management}. A challenge with these data is that the frequency of acquisition can be very variable, from a day to a few months depending on the agricultural practice. 

\subsection{Challenges}
\paragraphb{Data acquisition} Predicting a crop phenotype from a multi-modal dataset including environmental, genomics, and field management data requires collecting a very complex and expensive experimental dataset. On the field, soil sensors, weather stations, and logging field management variables are needed in a multi-site and multi-condition setup. Resequencing data from hundreds to thousands of lines are also needed in order to obtain a genomics dataset that captures the diversity of the organism. Due to the large complexity of acquiring such datasets, few have been collected to date. The G2F initiative is the most complete and extensive dataset that has been collected to date and is publicly available. Accordingly, and similar to previous studies, we only use these data in our current work.

\paragraphb{Data modeling} It is hard to model how each of the factors considered in this study (G, E, M) interact with each other and together affect the eventual phenotype. For instance, genomics data interactions (G*G), also called epistasis, are highly complex due to the existence of polysemy and distant semantic relationships, which are hard to capture ~\cite{gg-interaction}. These interactions, however, play a key role, since a gene's effect on the phenotype can be enhanced or diminished by other genes. In addition, different genes respond differently to environmental conditions, which makes very difficult to model interactions between genetic and environmental factors (G*E). Due to the complexity of these interactions, these factors are often modeled independently from each other \cite{washburn2021predicting}. However, it has been previously shown that the interaction between multiple factors can have a big impact on the phenotype \cite{kick2022yield}. For instance, variation in yield is largely controlled by interactions between genetic and environmental factors (G*E), rather than by genetic main effects alone \cite{rogers2021importance}, and the best genotype of sugarcane was shown not to be the best at another altitude \cite{sugarcane_alt_genome}.





\section{Dataset}
\label{sec:dataset}
\vspace{-0.3cm}
The G2F initiative~\cite{g2f}, which we use in our current work, contains data for 2000 unique maize seeds, sown in 32 different agricultural fields across 18 states, and was collected annually from 2014 to 2017. It originally contained approximately $50$k rows, but we used a filtered dataset of size $36$k, after removing rows with missing or inaccurate data, based on~\cite{anna-ge-dominance}. The filtered dataset has a total of $67$ unique location-year combinations, which differ in environmental, social, and technical conditions that influence crop growth. It also contains $1873$ unique seed hyrbids. Each data point in this dataset comprises of (i) genome of the seed variety, (ii) location, weather conditions and soil characteristics of the field, (iii) field management information (iv) yield data for fields (phenotype). 
We now delve into each of these data types.

\paragraphb{Genome} Each datapoint includes the values for 955,690 SNPs for the genome, based on reference genome Zea mays V2~\cite{zea-mays-genome}. The SNPs were then cleaned and filtered to remove those with missing data, based on methods listed in ~\cite{anna-ge-dominance}, reducing the number of SNPs to 20372. As described in Section~\ref{sec:framework-dna}, we select 100 SNPs out of these based on mutual information.

\paragraphb{Weather} This dataset includes weather data taken from the Spectrum Watchdog weather stations located at each field station described previously~\cite{anna-ge-dominance}. However, due to missing data and potential errors, we chose to crawl weather data from Daymet~\cite{daymet-weather} based on each field's location. We collected values for six features, including solar radiation, vapor pressure, precipitation, maximum and minimum temperature and wind speed for every day in the planting season. We added three more features -- dew point, relative humidity and growing degree days, which are known to be correlated with grain yield, based on the formulae given in ~\cite{dew-point-formula, corn-gdd-formula}. We have a total of nine weather-based features. We averaged each of these over 5-day windows to obtain a time-series of length 43 for each feature.

\paragraphb{Soil} The dataset includes information on a total of 19 features that measure physical and chemical properties of soil. These include pH, Calcium \& Magnesium content, and organic matter based on soil samples taken from some of the fields in 2016 and 2017. We used SSURGO dataset to fill in the missing values. 

\paragraphb{Field Management} Since field management data was only partially available across the four years, we just used five coarse-grained features, i.e. aggregate irrigation, total fertilizers and planting density based on the number of seeds sown.
\section{Framework}
\label{sec:framework}
\begin{figure}
    \centering
    \includegraphics[width=0.5\textwidth, height=1
    in]{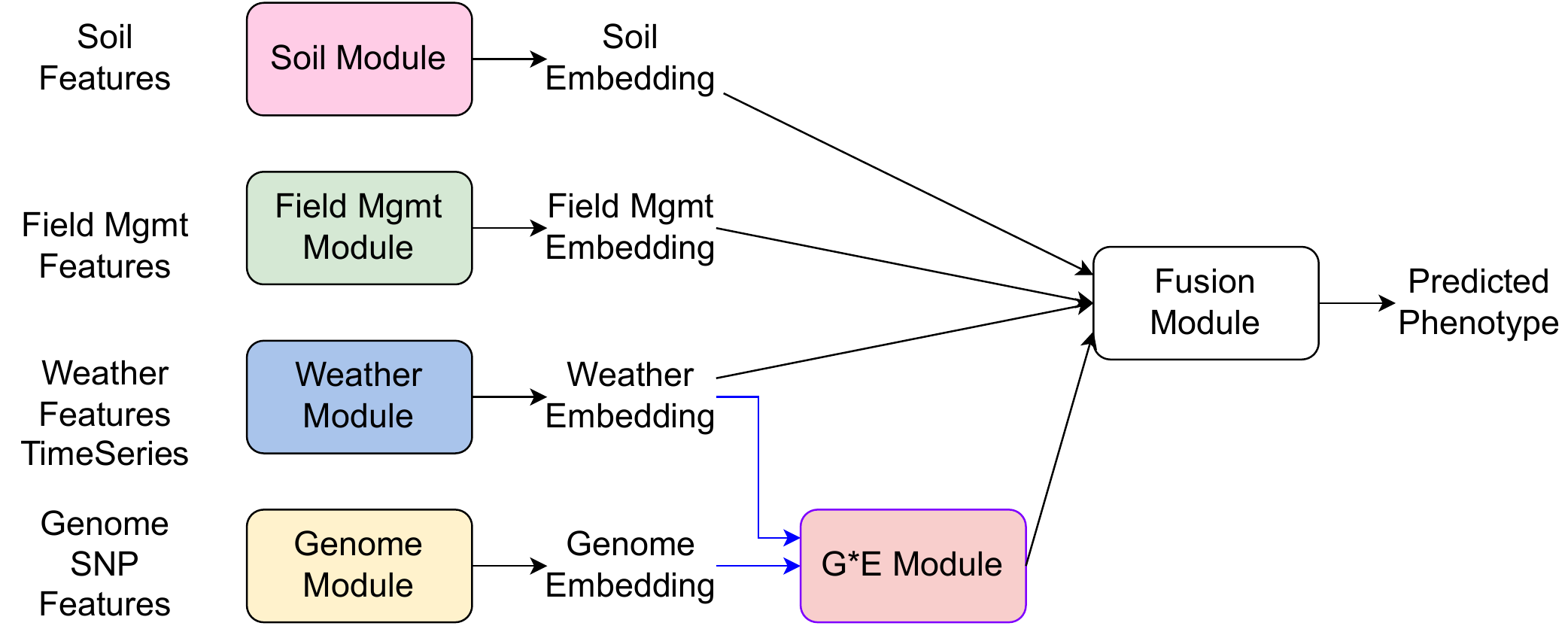}
    \caption{Overall architecture capturing the G, E and M interactions.}
    \label{fig:overall_architecture}
\end{figure}

The goal of our framework is to predict grain yield using multi-modal data including genomics (G), environment (E) i.e. weather and soil, and field management practices (M).
Figure~\ref{fig:overall_architecture} illustrates how our model uses separate modules to generate an embedding for the genome (SNPs), weather, soil and field management features. We augment the G embedding with a cross-attention based G*E module. We then concatenate all the embeddings and send them to a fusion module, which outputs the predicted grain yield. We now describe all the components in detail.

\subsection{Modeling DNA}\label{sec:framework-dna}
 
We represent each SNP with a one-hot encoding of length 4, denoting the four possible nucleotides (A, C, G and T). Heterozygous SNPs, i.e., SNPs that take two different values (for the two alleles for the case of diploid organisms), are denoted as having half of two bases. For instance, the degenerated nucleotide k is represented as 0.5 G and 0.5 T. 

\paragraphb{Feature Selection} To ensure computational tractability, we use feature selection methods to only consider the SNPs with the highest predictive power for grain yield. We first use Recursive Feature Elimination~\cite{rfe-sklearn} to pick the top 1000 SNPs, by iteratively training a LightGBM model~\cite{lightgbm} to predict grain yield using just the SNPs. Next, we select 100 SNPs with the highest mutual information~\cite{mutual-info-sklearn} with respect to grain yield. 

\begin{figure}
    \centering
    \begin{minipage}{0.5\textwidth}

    \includegraphics[width=\textwidth]{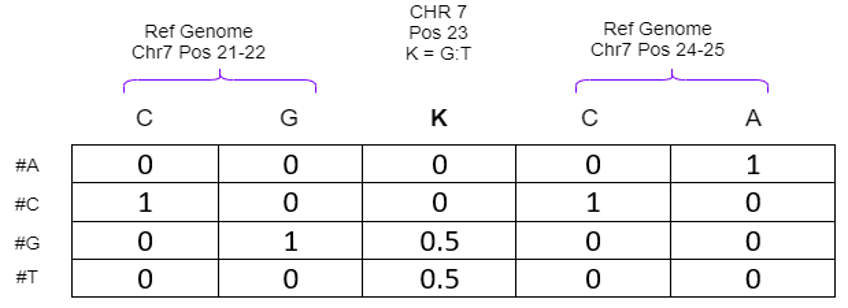}
    \caption{We append neighboring DNA sequence from the maize reference genome. In the above example, there is a SNP at chromosome 7 and position 23. This particular seed is a hybrid with G on one strand and T on the other, since its parent seeds have G and T, respectively at this SNP. The two bases on left and right are taken from the reference genome positions 21-22 and 24-25, respectively. Then the sequence is one hot encoded, with K being 0, 0, 0.5, 0.5. }
    \label{fig:context_dna}
    \end{minipage}


\end{figure}

\paragraphb{Modeling Genome-to-Genome (G*G) interactions}
\label{para:g_g}
We model DNA as a natural language since this allows us to capture complex  interactions. We leverage the fact that DNA sub-sequences (k-mers\footnote{k-mers are sequences of DNA of length k.}) together with the neighboring k-mers decide their effect on a phenotype. First, we obtain more context around a SNP. We achieve this by appending nearby DNA sequences from the reference genome (around the SNP position). For example, in Figure~\ref{fig:context_dna}, we append 2 bases on both sides of the SNP, and each base in the resulting 5-base sequence is represented using one-hot encoding. Second, we use a one-dimensional convolutional neural network (CNN) over each SNP sequence with  multiple kernels of lengths 2, 3 and 4 to capture G*G interactions. We then use max-pooling to get an embedding for each SNP. Note that we use the same set of kernels for all SNPs. 
Next, we add positional information to each SNP embedding by using sin-cos functions to encode which chromosome, and which position within the chromosome, the SNP represents. This is similar to positional encoding used in~\cite{transformer}. 

\paragraphb{Modeling G*E interactions}
To model a second type of interactions, which comes from environmental factors such as weather, soil and field management practices data (G*E), we utilize the fact that same k-mer can affect a phenotype differently under different environments. More specifically, a phenotype is influenced by the combined interactions of all k-mers in a given environment. To capture the G*E interaction, specifically to add weather as context to the SNP, we employ a cross-attention module based on~\cite{tan2019lxmert}. For each SNP, we treat the SNP's embedding (as output by the Genome Module described in Para~\ref{para:g_g}) as the query vector $x$, and the weather embedding sequence,  $\{y_1, y_2, \dots, y_N\}$ to calculate the matching score $\alpha_i$ between the query vector and each context vector, i.e.
\[ \{\alpha_i \} = \text{SoftMax}( \{ \text{score}(x,y_1),\ldots,\text{score}(x,y_N) \} ) \]

Intuitively, the matching score $\alpha_i$ between the SNP embedding and each context vector signifies how important is each timestep of the weather features with respect to the SNP. We use single-head attention to get the weighted sum of context vectors as the output, i.e. $\text{Att}(x,\{y_i\}) = \sum_j \alpha_j y_j$.
We then add a single dense layer to get the final G*E embedding of the same length as the SNP embedding. We then add its output to the SNP's embedding. Lastly, we have a max pooling layer to combine the embeddings of all the SNPs into a single genome embedding.


\section{Modeling environment}
\paragraphb{Modeling Weather Features}\label{ref:framework-weather}
To capture temporal context from the time series, we use a CNN~\cite{cnn-for-time-series}, wherein each weather feature is modeled as a separate channel. We use a series of one-dimensional convolutional layers followed by a max pool layer to get an embedding sequence for the weather features.

\paragraphb{Modeling Soil and Field Management Features}
We process the soil features with a series of fully connected layers with ReLU to get an embedding capturing all the soil features.
For coarse-grained field management features, we use a series of fully connected layers with ReLU to get a vector embedding for the 

\begin{figure}
    \centering
    \includegraphics[width=0.4\textwidth]{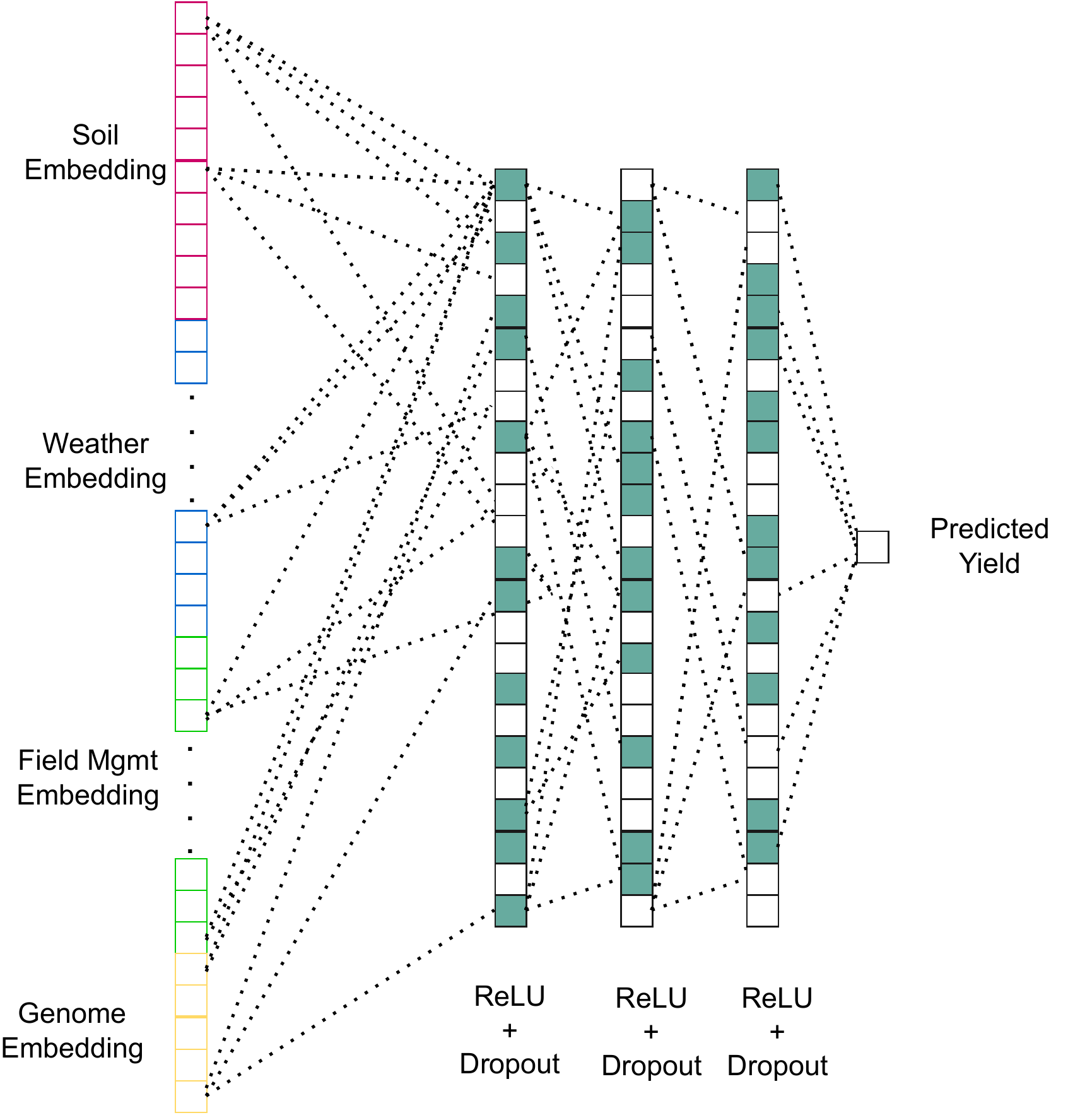}
    \caption{Architecture of the Fusion Module: we flatten and concatenate the embeddings of all input features and pass them through a series of fully connected layers with ReLU activation and dropouts.}
    \label{fig:fusion-module}
\end{figure}
\section{Fusion Module}
Figure \ref{fig:fusion-module} shows the architecture of the last module in our framework, which fuses the embeddings of the genome, weather, soil and field management together, and applies a series of fully connected layers with ReLU and dropouts. It then outputs the predicted phenotype.
\section{Evaluation}
\label{sec:eval}
\subsection{Setup} 
\begin{figure*}
    \centering
    \begin{subfigure}[b]{0.4\textwidth}
    \includegraphics[width=\textwidth]{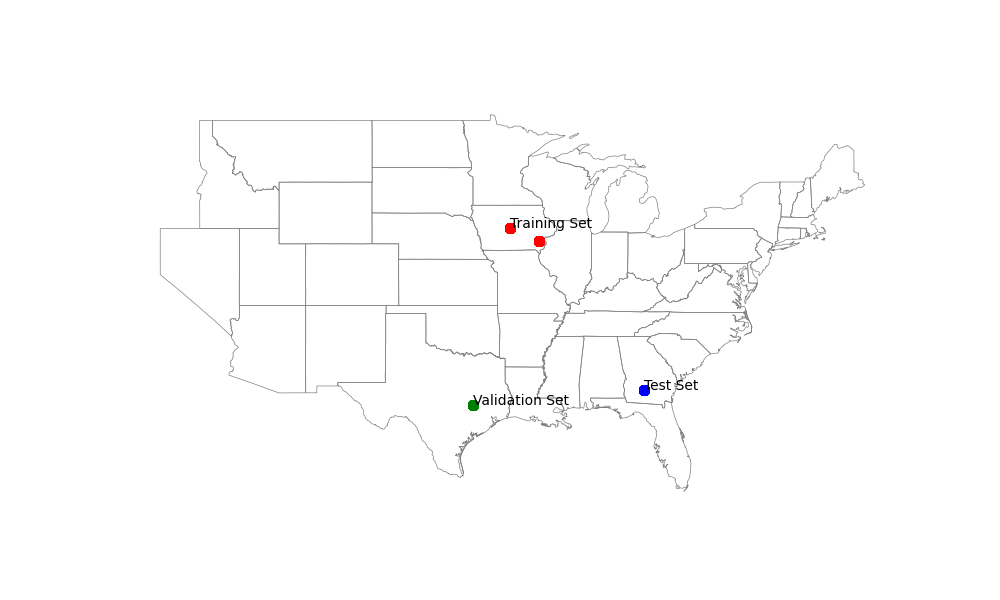}
    \caption{Environment Split}
    \label{fig:environment_split}
    \end{subfigure}
    \begin{subfigure}[b]{0.4\textwidth}
    \includegraphics[width=\textwidth]{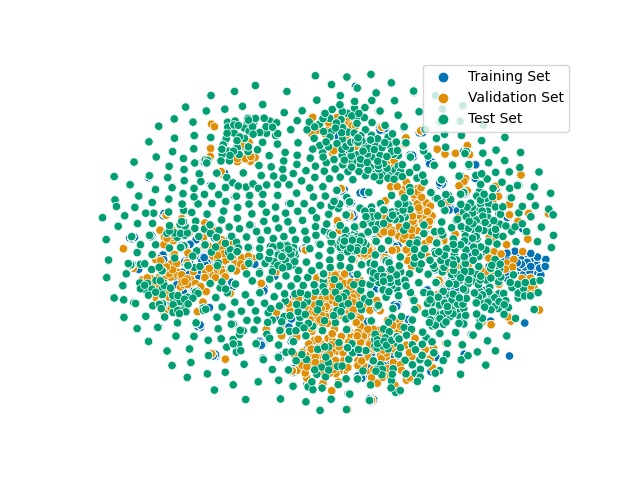}
    \caption{t-SNE plot for the hybrid split}
    \label{fig:hybrid_split}
    \end{subfigure}
\end{figure*}

\textbf{Data Splits}. To evaluate our framework, we consider two scenarios, (i) Environment split: Here, we randomly selected location-year pairs to be included in the test and validation datasets, and these were completely excluded from the training dataset. This scenario tests how well our model generalizes to environments unseen, (ii) Hybrid split: In this case, we first used K-Means clustering to aggregate the 2000 seeds into 100 clusters, and then randomly assigned clusters to test and validation datasets, and used all the data points for seeds in the remaining clusters as the training dataset. This scenario tests if our model generalizes to new hybrids. 
We show the environment and the hybrid split in Figures ~\ref{fig:environment_split} and ~\ref{fig:hybrid_split} and . The environment split in Figure~\ref{fig:environment_split} shows that the training, validation and test locations form non-overlapping sets. On the other hand, for the hybrid split (Figure~\ref{fig:hybrid_split}) there is overlap in some places for the three sets. For visualization purposes, we used t-SNE~\cite{tsne} dimensionality reduction that was initialized with PCA. This reduced a dimension 100 genome vector to dimension 2 for visualization. The observed overlap is expected because the SNPs are captured from a population of hybrid seeds, so partial copies of their genomes should be found through the population. Since, in addition, the SNPs are derived from a finite set of nucleotides, the overlap in the SNPs should be present by design. 

For cross-validation, we split the training dataset into \numfolds folds, out of which 8 folds form the training set, and the rest two form test and validation set respectively. 

\textbf{Metrics}.
We use Pearson correlation coefficient and Root Mean Squared Error (RMSE) as the main metrics to compare the predicted and ground truth phenotype values. We choose Pearson correlation coefficient because this is also a metric used by ~\cite{Washburn2021-gem-cnn}, which is one of our closest baselines. We report the mean and standard deviation of both the metrics across \numfolds folds.

\textbf{Data Preprocessing}. The length of the weather time-series was unequal across different data points because of differences in when the crops were planted and harvested. We therefore padded the time series for each feature with the average value of that feature. Also, we use standard scaling~\cite{sklearn-standard-scaler} to normalize all the features that are input to the model.

\textbf{Hyperparameter Tuning}.
We used hyperopt to perform hyper-parameter tuning for the model (for parameters such as number of convolutional layers). Due to limited resources, we optimized the parameters only for the Environment-split scenario,
and used the same hyper-parameters for all of our evaluation.


\textbf{Baselines}. We compare DeepG2P to a statistical model, GEBlup model~\cite{henderson1975best} available with the R package qgg, a mechanistic model, AutoCGM (ApSim simulator)~\cite{holzworth2014apsim,holzworth2018apsim} and CNN-21, another fusion based deep learning model~\cite{Washburn2021-gem-cnn}. The setup and parameters are as described in ~\cite{Washburn2021-gem-cnn}. 

\subsection{Comparison with Baselines}
\begin{figure*}
    \centering
    \begin{subfigure}[b]{0.3\textwidth}
    \includegraphics[width=\textwidth]{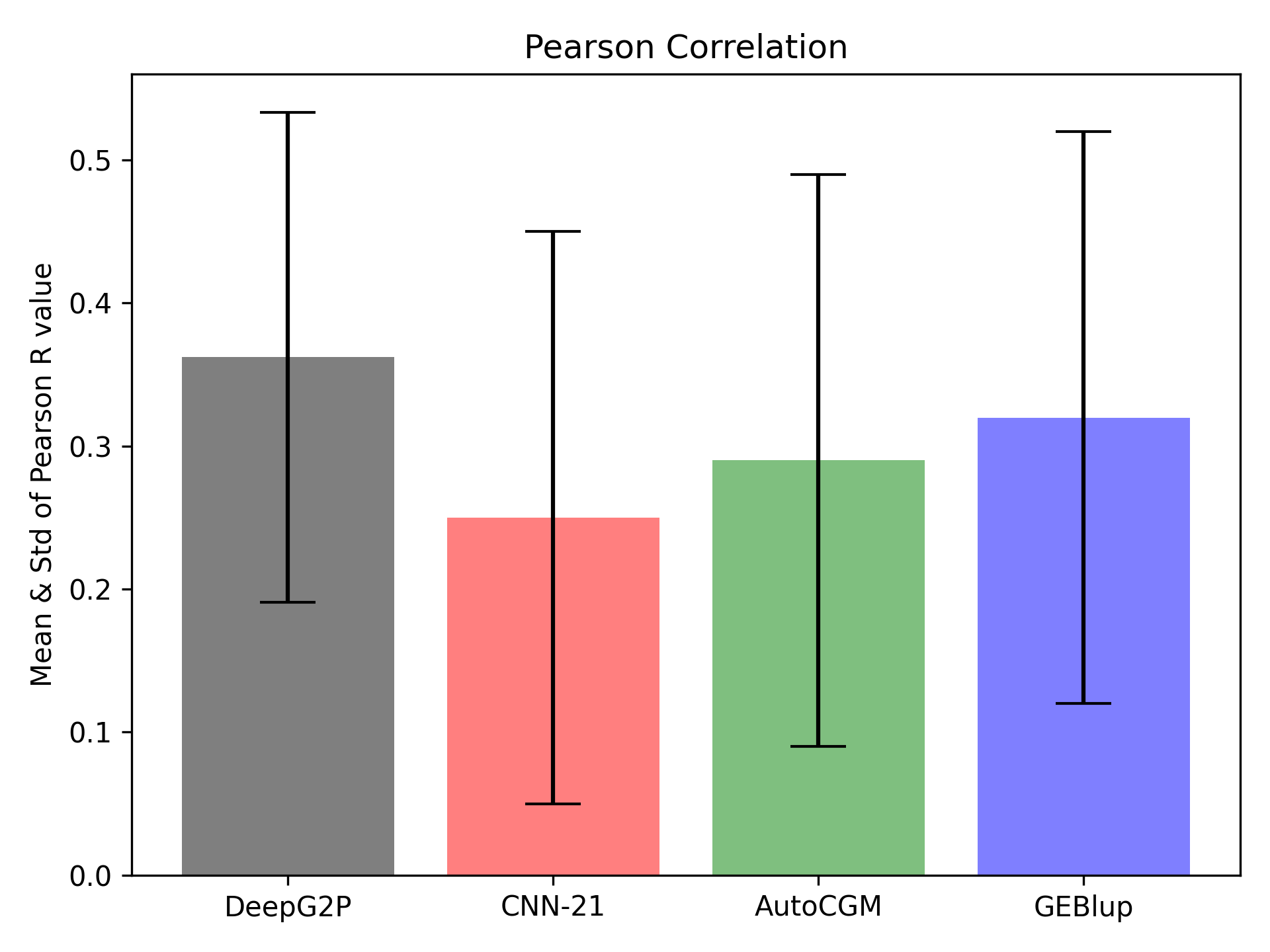}
    \caption{Pearson Correlation Coefficient}
    \label{fig:pearson_envt_split}
    \end{subfigure}%
   \begin{subfigure}[b]{0.3\textwidth}
    \includegraphics[width=\textwidth]{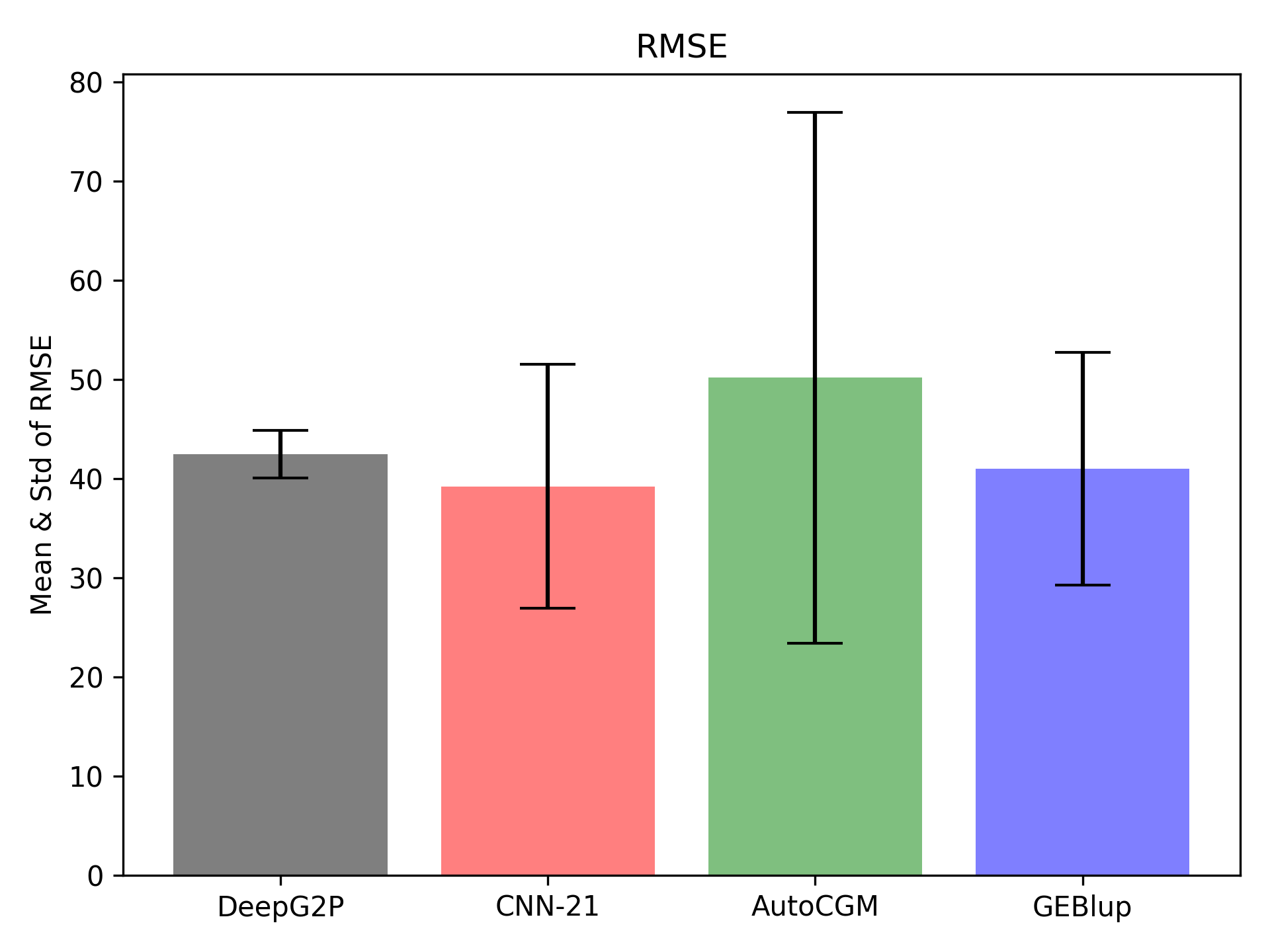}
    \caption{RMSE}
    \label{fig:rmse_envt_split}
    \end{subfigure}
    \caption{Comparison with baselines for the environment split}
    \label{fig:envt_split}
    \begin{subfigure}{0.3\textwidth}
\includegraphics[width=\textwidth]{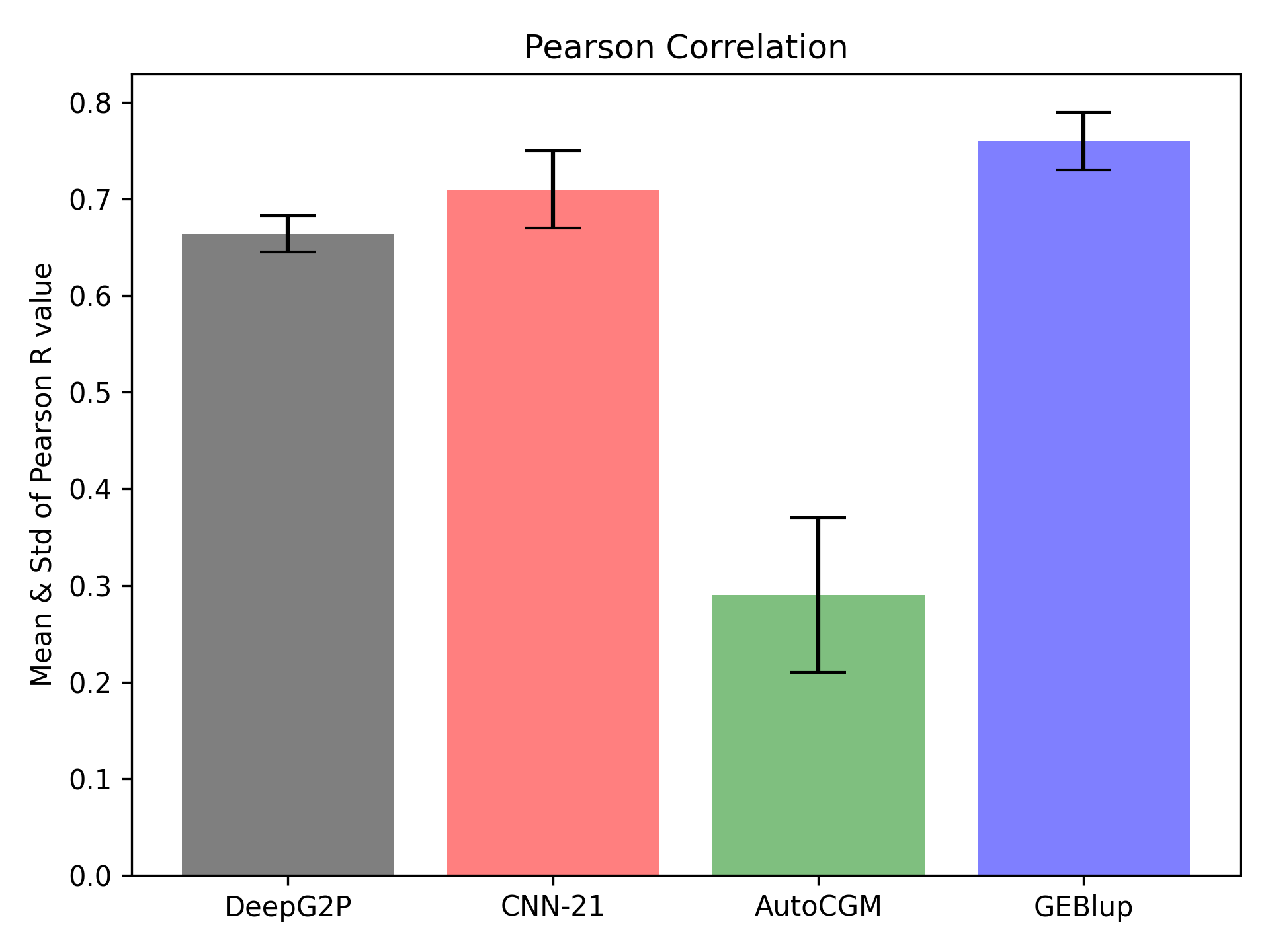}
    \caption{Pearson Correlation Coefficient}
    \label{fig:pearson_hybrid_split}
    \end{subfigure}%
    \begin{subfigure}{0.3\textwidth}
    \includegraphics[width=\textwidth]{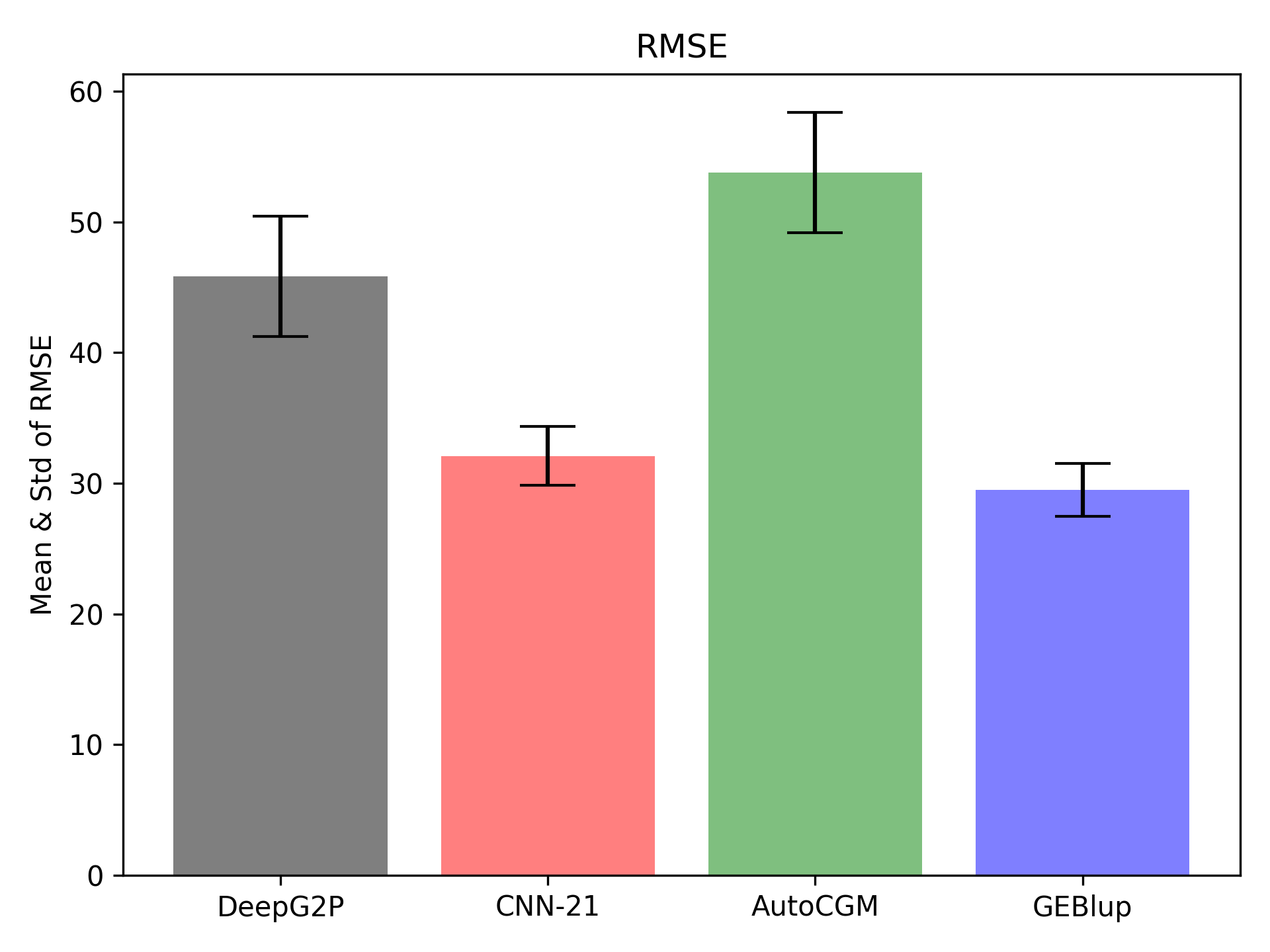} 
    \caption{RMSE}
    \label{fig:rmse_hybrid_split}
    \end{subfigure}
 \caption{Comparison with baselines for the hybrid split}
 \label{fig:hybrid_split}
\end{figure*}

In this section, we present results comparing the performance of DeepG2P with the other three baselines. Figure~\ref{fig:envt_split} shows our results for the environment split. We observe that our approach outperforms all the approaches in terms of both the Pearson correlation coefficient and RMSE, which indicates good generalization power to new unseen locations. Results show that Pearson's correlation coefficient is $1.45\times$ better. RMSE, on the other hand, is worse by $8$\%, but with lower variance than CNN-21. One of the reasons for this could be that our approach uses the G*E cross-attention layer that allows us for each SNP to pay attention to the environment, that is not present in the CNN-21 approach. It is also interesting to note that GEBlup outperforms the other deep learning approach CNN-21. Note that GEBlup explicitly utilizes the interactions between genome and weather as well as genome and soil, respectively. The mechanistic model AutoCGM does better than CNN-21 in this case. Figure~\ref{fig:hybrid_split} reveals a different pattern. Barring AutoCGM, our approach performs similarly, although slightly worse than than CNN-21 and GEBlup in terms of the Pearson correlation coefficient and RMSE. It should be noted that we only retain 100 SNPs. However, the CNN-21 approach uses PCA for dimensionality reduction over all the SNPs. GEBlup utilizes all the 20k SNPs as it requires only linear calculations, without running into computational issues. When using the hybrid split, the distinguishing feature of the hybrids is the genomic content and using more SNPs i.e., more genomic content leads to better results. However, DeepG2P does not do much worse indicating that careful selection of SNPs can lead to much better results. Comparing the results of DeepG2P over the two types of data splits, we can note the superior performance when encountering new locations, which is difficult for other approaches to achieve.

\subsection{Ablation Study}
\begin{figure*}
    \centering
    \begin{subfigure}{0.3\textwidth}
    \includegraphics[width=\textwidth]{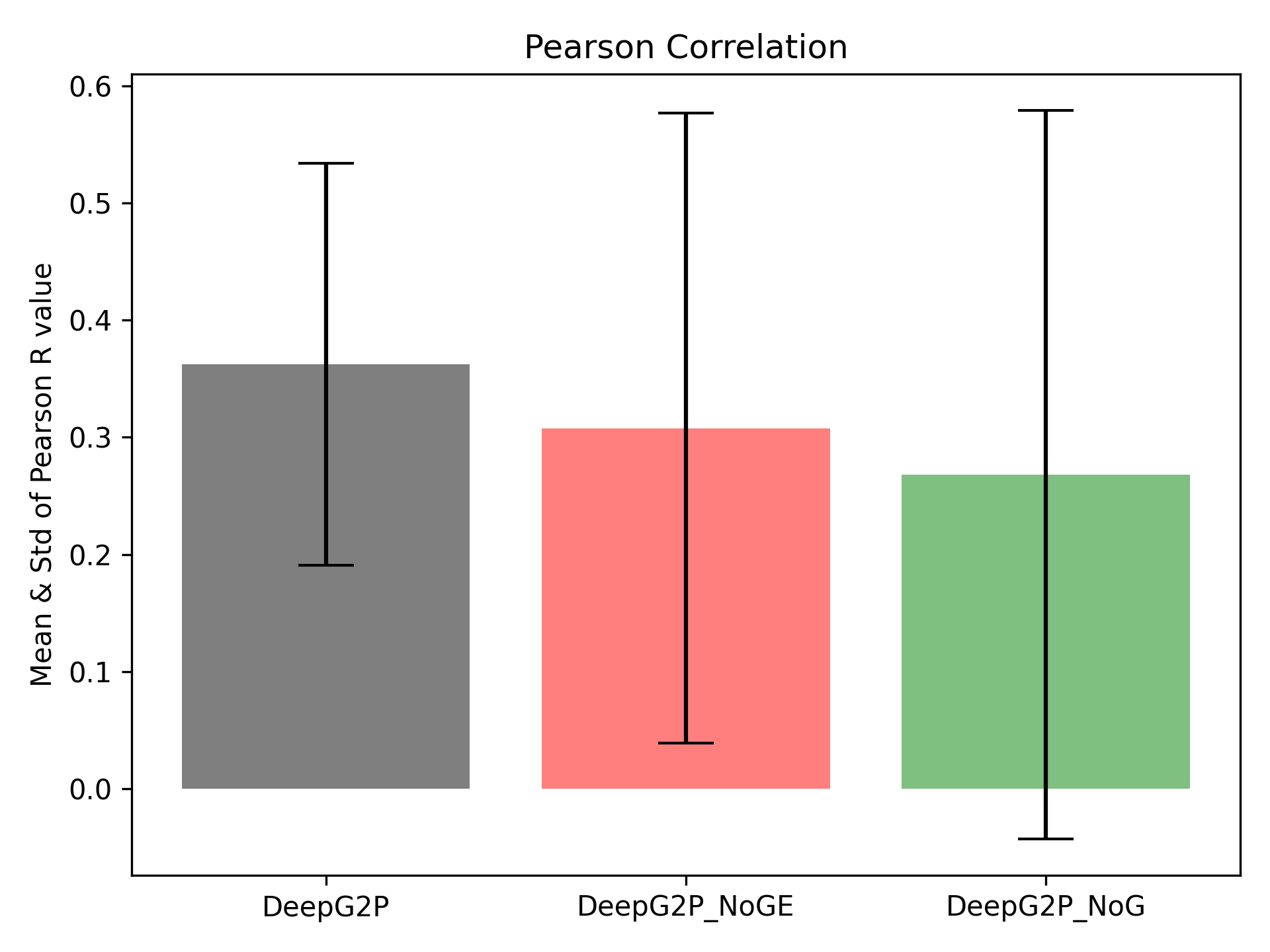}
    \caption{Pearson correlation coefficient}
    \label{fig:ablation_pearson_envt_split}
    \end{subfigure}%
    \begin{subfigure}{0.3\textwidth}
    \includegraphics[width=\textwidth]{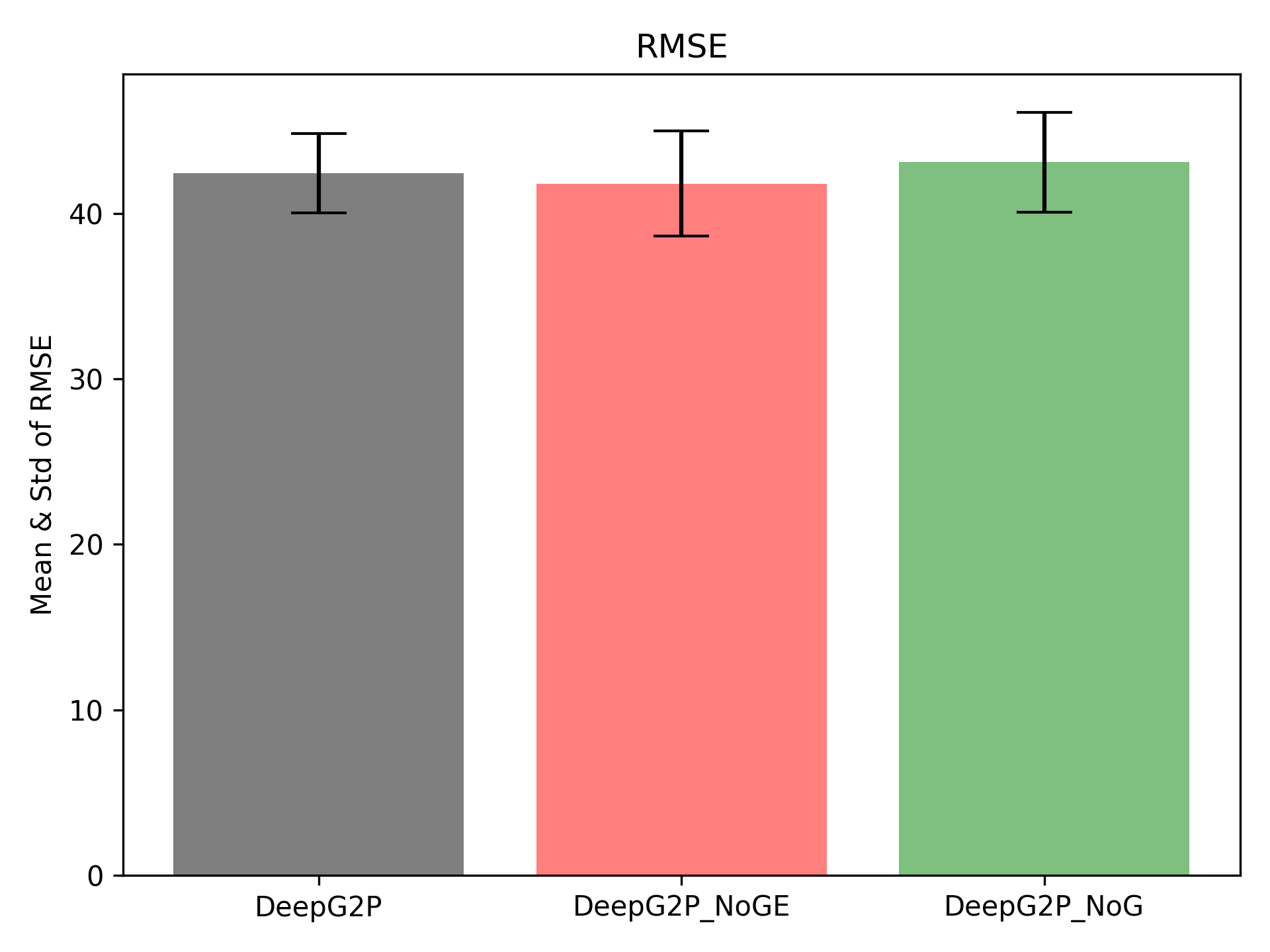}
    \caption{RMSE}
    \label{fig:ablation_pearson_envt_split}
    \end{subfigure}
\caption{Results of ablation study for the environment split}
\label{fig:ablation_env}
\end{figure*}

We also carried about an ablation study in order to study the effect of the genome and G*E cross-attention modules. We compared vanilla DeepG2P with DeepG2P\_NoGE by removing the GE cross-attention, and with DeepG2P\_NoG by removing all SNP related columns i.e., removing all genomic content. The results in Figure~\ref{fig:ablation_env}(a-b) indicate that for the environment split the Pearson correlation coefficient is worse by 15\%, however the RMSE is similar. An increased variance is observed for both metrics. It is also seen that genomic data is critical to the performance of DeepG2P. In contrast to the environment split, not much performance degradation is seen in the case of the hybrid split. A possible reason is that variations in environment are more pronounced than the variations in the genome for the SNPs we selected. As a result, the model's learning relies mostly on the environment related variables.    

\begin{figure*}
    \centering
    \begin{subfigure}{0.3\textwidth}
    \includegraphics[width=\textwidth]{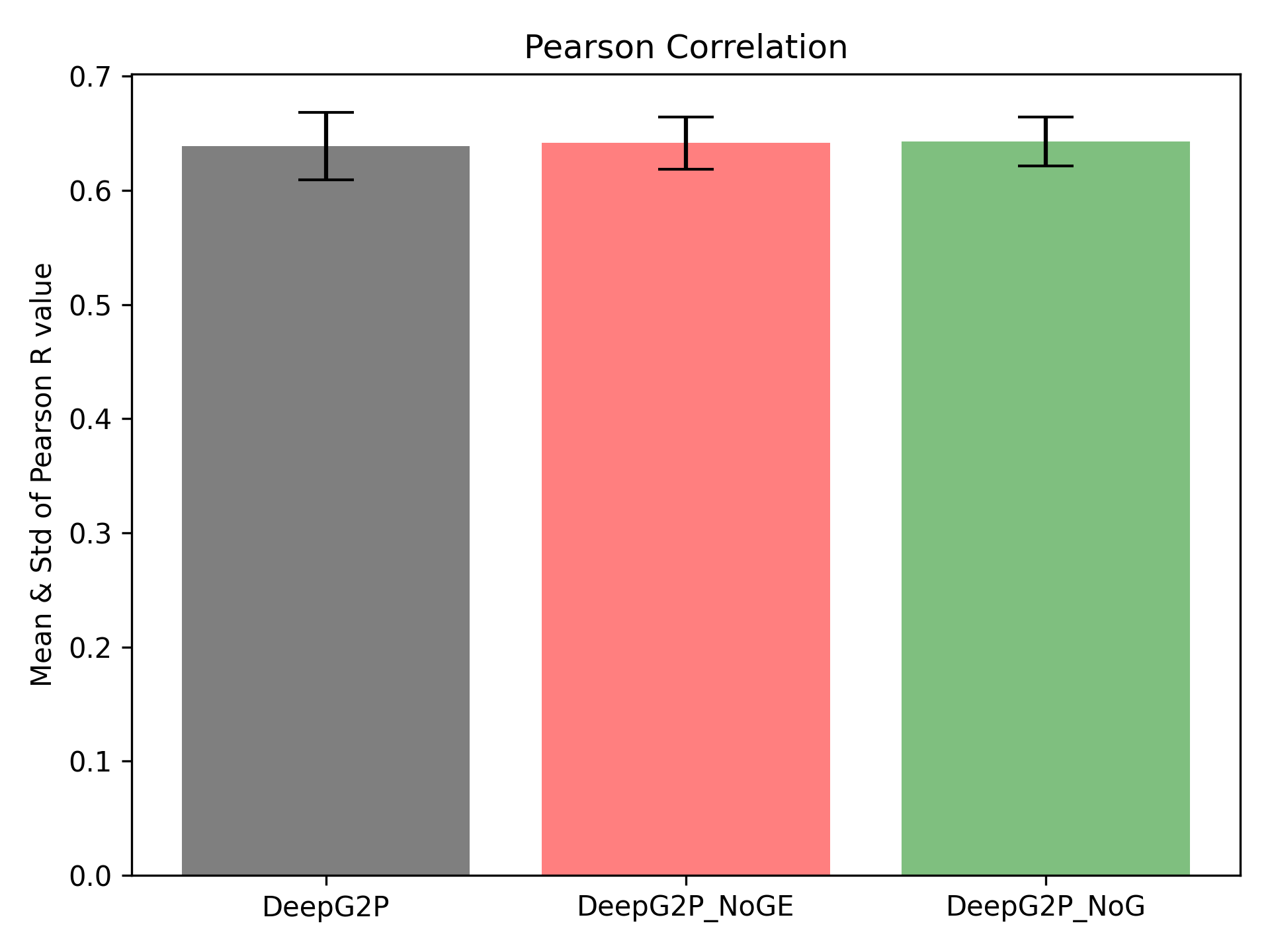}     \caption{Pearson correlation coefficient}
    \label{fig:ablation_pearson_hybrid_split}
    \end{subfigure}%
    \begin{subfigure}{0.3\textwidth}
    \includegraphics[width=\textwidth]{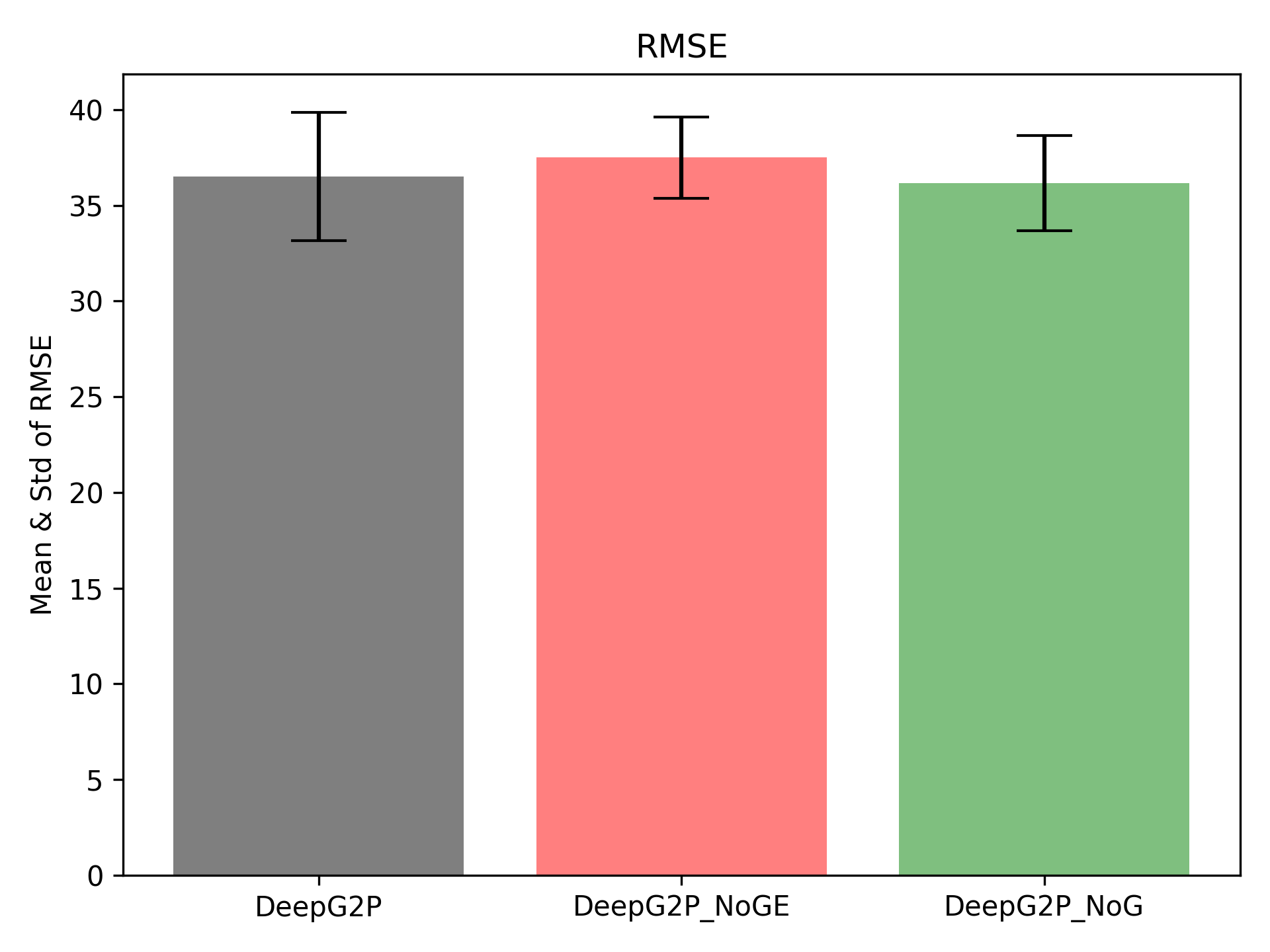} 
    \caption{RMSE}
    \label{fig:ablation_pearson_hybrid_split}
    \end{subfigure}
    \caption{Results of ablation study for the hybrid split}
\label{fig:ablation_hybrid}

\end{figure*}
\section{Discussion and Future Work}
\label{sec:discussion}
In this work we presented a novel deep learning approach to fuse multiple modalities to predict crop yield. Our proposed architecture processes genomics, environmental, and field management data in separate modules, but unlike some of the work that had been previously done in this space~\cite{washburn2021predicting}, it also models interactions across data types. Specifically, by making use of a G*E cross-attention module, we tackle how different genotypes react differently to the changing environment. This is a key addition, since it is well accepted that the interaction between genome and environment plays an important role in phenotypic outcome. 
This addition has resulted in a strong performance of our model. Specifically, our results show that our approach outperforms other mechanistic crop growth models (AutoCGM), statistical approaches (GEBlup) and deep learning models (CNN-21) for unseen environments and does similarly to these approaches for the unseen seed varieties.

We also show that each approach performs worse in the environment split than it does in the hybrid split. This suggests that the environmental data is too limited and indicates a need for more such datasets, with more locations for training such that the i.i.d. assumption is not violated. Although it is well understood that the environment and  phenotype have a causal relationship (e.g., more sunlight may lead to better quality agricultural produce); it is possible that the environments may not come from a stationary distribution and thus deep learning approaches can't properly generalize. 

Similarly, our architecture models genomic interactions (G*G) to capture the fact that the effect of a SNP (or gene) on the phenotype can be enhanced or diminished by other SNP (or gene). For that, we proposed a novel approach to get SNP context from the reference genome and modeled local interactions in the genome using CNNs. The genomic data, however, presents additional challenges. Specifically, the number of features (SNPs) tends to be orders of magnitude larger than the number of observations (genotypes or cultivars) so applying dimensionality reduction techniques to the features is a typical practice. However, the ability, not only to predict a phenotype, but also to determine which SNPs are causal of the phenotype, is a recurring and challenging question that plant scientists and breeders face. 
By maintaining the SNP dimensionality space, we aimed at ensuring that these results can be utilized by downstream field applications. 
Although this is a very important feature of our approach, it also exposes new challenges associated to processing $>$20k SNPs. To address this challenge, we used a two step approach to select the top 100 most influential SNPs for the phenotype. 

Although using 100 SNPs still led to a good performance of our approach, it is unquestionable that using all 20k SNPs would provide more resolution on each of the genotypes for training our model. Therefore, we hypothesize that this reduction of the number of SNPs used in the prediction is causing a decrease in the performance of the model on the hybrid split as compared to the environmental split.  Specifically, the baseline models, which use all SNPs or apply dimensionality reduction, perform better in the hybrid split. This is an expected behaviour, since, as mentioned above, the environmental split is a more challenging setup than the hybrid split. However, our approach performs worse in the hybrid split. This suggests that an increase in the number of SNPs processed by our method could further improve our performance. To test this hypothesis, our next step will focus on using recent advances in GPU computing to process all the SNPs. In parallel, it will be interesting to study how SNP selection can be done for multiple phenotypes. Finally, another direction of our future work will focus on testing our approach on other crops. 
\vspace{-0.25cm}  
\subsection{Societal implications} 
By 2050, the world population will reach between 8.3 and 10.9 billion people, and such growth rates will require an increase in food supply of 50\% - 75\%. Meeting rising food demand in the context of global warming will also require an understanding of how agricultural output responds to climatic variability ~\cite{lesk_influence_2016}. Farmers need a better knowledge of the effects of climate on agricultural productivity~\cite{zampieri_wheat_2017}. Precision agriculture, which uses emerging technologies to help increase crop yields and improve efficiency, will play an essential role in this effort. Specifically, in PA, machine learning models can process field inputs to make field management decisions. Genomics data, however, has not traditionally been included in the PA schema. In this context, our work aims to fill a gap in the AI solutions built for PA. The use, not only of field measurements inputs, but also the genomic background of the cultivars, can boost the performance of these models, and therefore improve yield by means of using the correct cultivars (with the predicted genomic backgrounds) in different locations or weather conditions. 

In addition, genomic data has been traditionally used for phenotypic predictions by breeders, using mechanistic or statistical methods. Breeders are in constant needs of generating new varieties that best respond to some environmental conditions or to new pathogen threads. Understanding the SNPs that lead to certain phenotypes is therefore an important question that can have an impact in food production. Our model also seeks to fill a gap in this context, where AI algorithms are much less explored, and statistical approaches continue to be the leading computational approach.

In summary, applied to real-world datasets, our model has the potential of helping farmers and breeders achieve better crop yield in changing environments.



\bibliographystyle{unsrt}
\bibliography{paper}

\end{document}


%

%

\onecolumn
\aistatstitle{DeepG2P: Fusing Multi-Modal Data to Improve Crop Production: \\
Supplementary Materials}



\section{Genomic variant (SNP) selection}
\begin{figure*}[h]
    \centering
    \includegraphics[width=0.5\textwidth]{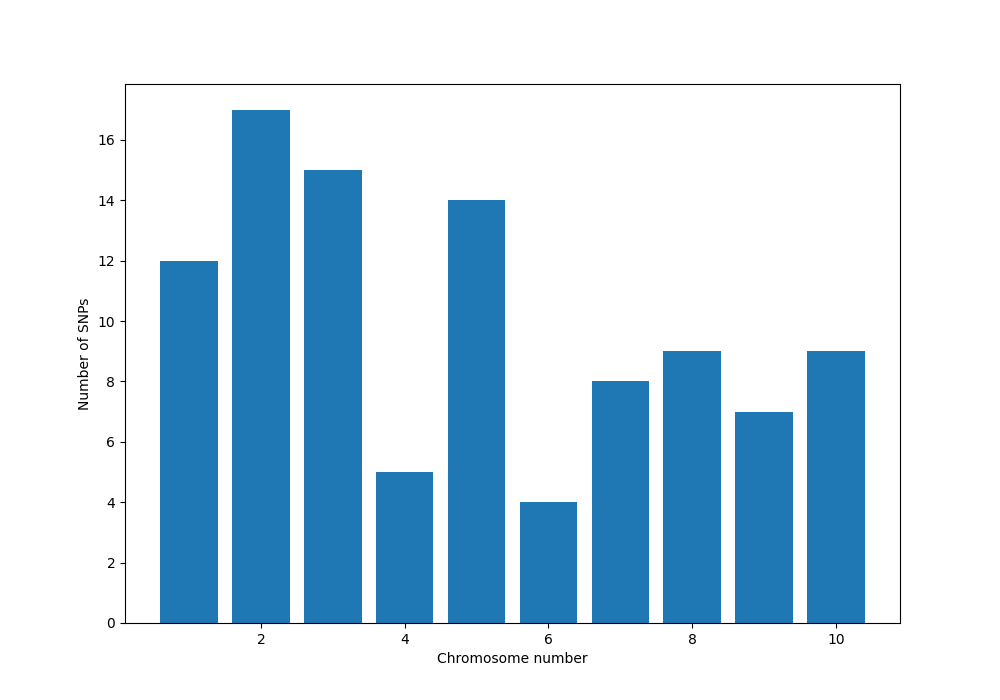}
    \caption{Distribution of SNPs over chromosomes}
\label{fig:snp_to_chr}
\end{figure*}

Figure~\ref{fig:snp_to_chr} shows the distribution of SNPs over chromosomes. Maize has ten chromosomes and we can see that the 100 SNPs we select are distributed over all the chromosomes. 

We also show LGBM split, gain, and mutual information gain for the top selected SNPs in Figures~\ref{fig:lgbm_split}, ~\ref{fig:lgbm_gain} and ~\ref{fig:MI}.
\begin{figure*}[h]
    \centering
    \begin{subfigure}[b]{0.3\textwidth}
    \includegraphics[width=\textwidth]{figures/supplementary/LGBM Split Importance of selected SNPs_top100.png}
    \caption{LGBM Split Importance for the top 100 SNPs}
    \label{fig:lgbm_split}
    \end{subfigure}%
    \begin{subfigure}[b]{0.3\textwidth}
    \includegraphics[width=\textwidth]{figures/supplementary/LGBM Gain Importance of selected SNPs_top100.png}
    \caption{LGBM Gain Importance for the top 100 SNPs}
    \label{fig:lgbm_gain}
    \end{subfigure}%
    \begin{subfigure}[b]{0.3\textwidth}
    \includegraphics[width=\textwidth]{figures/supplementary/Mutual Info Gain bw SNPs and Grain Yield_top100.png}
    \caption{Mutual Information Gain between the top 100 SNPs and Grain Yield}
    \label{fig:MI}
    \end{subfigure}
    
\end{figure*}
\section{Architecture choices for each module}
\begin{figure*}[h]
    \centering
    \begin{subfigure}[b]{0.5\textwidth}
    \includegraphics[width=\textwidth]{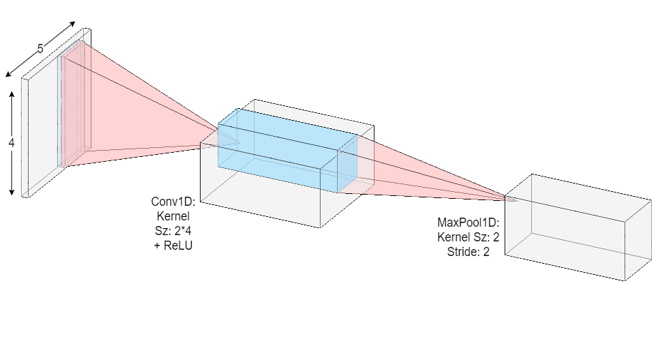}
    \caption{Genome module architecture. Each SNP is of dimension 4 by 5, where 4 is the length of the one-hot-encoded vector for each base pair and 5 is the length of the context around the SNP position and the SNP itself.}
    \label{fig:genome}
    \end{subfigure}%
   \begin{subfigure}[b]{0.5\textwidth}
    \includegraphics[width=\textwidth]{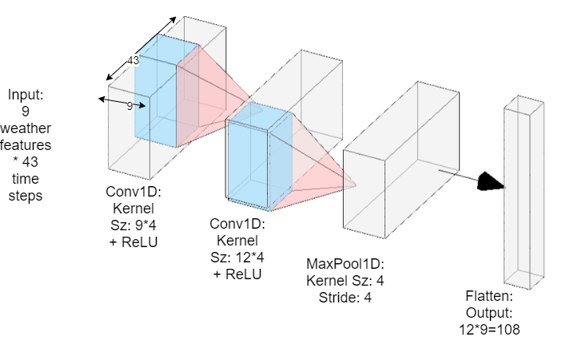}
    \caption{Weather module architecture}
    \label{fig:weather}
    \end{subfigure}
    \begin{subfigure}[b]{0.5\textwidth}
    \includegraphics[width=\textwidth]{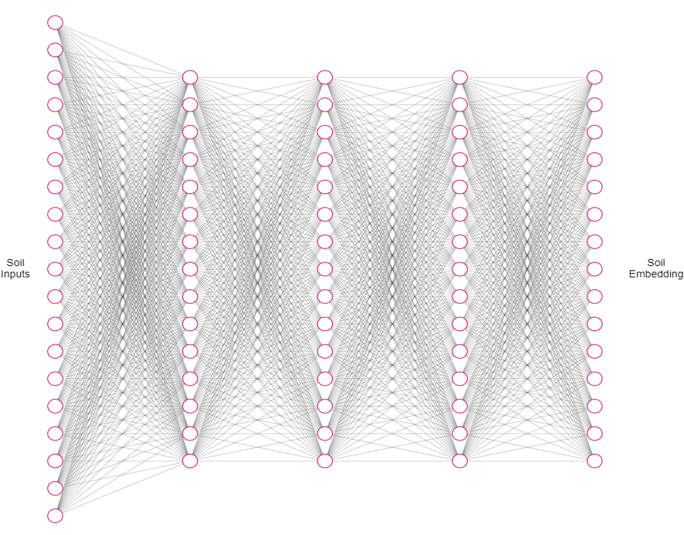}
    \caption{Soil feature architecture}
    \label{fig:soil}
    \end{subfigure}%
    \begin{subfigure}[b]{0.5\textwidth}
    \includegraphics[width=\textwidth]{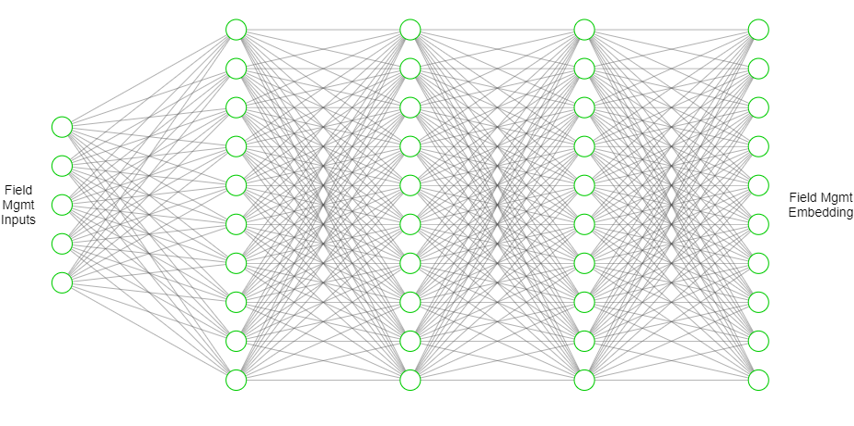} 
    \caption{Field Management architecture}
    \label{fig:field_mgmt}
    \end{subfigure}
\end{figure*}

We present the architectural details of each module. The genome module in shown in Figure~\ref{fig:genome}, the weather module in Figure~\ref{fig:weather}, the soil module in Figure~\ref{fig:soil} and field management module in Figure~\ref{fig:field_mgmt}. 

